%% file: main.tex
\documentclass[10pt, a4paper]{article}
\usepackage{lrec}
\usepackage{multibib}
\newcites{languageresource}{Language Resources}
\usepackage{tabularx}

\usepackage{epstopdf}
\usepackage[utf8]{inputenc}

\usepackage{hyperref}
\usepackage{xstring}

\usepackage{adjustbox}
\usepackage[nolist]{acronym}
\usepackage{graphicx}
\usepackage{rotating}
\usepackage{pdflscape}
\usepackage{hyperref} 
\usepackage{amsmath}
\usepackage[T1]{fontenc}
\usepackage{csquotes}
\usepackage[normalem]{ulem}
\usepackage[justification=centering]{caption}
\usepackage{hyperref}
\usepackage{xspace}
\usepackage{xargs}                      
\usepackage[pdftex,dvipsnames]{}  
\usepackage{pdfpages}
\usepackage[colorinlistoftodos,prependcaption,textsize=tiny]{todonotes}
\usepackage{color,soul}
\usepackage{caption}
\usepackage{subcaption}

\newcommandx{\unsure}[2][1=]{\todo[linecolor=red,backgroundcolor=red!25,bordercolor=red,#1]{#2}}
\newcommandx{\change}[2][1=]{\todo[linecolor=blue,backgroundcolor=blue!25,bordercolor=blue,#1]{#2}}
\newcommandx{\info}[2][1=]{\todo[linecolor=OliveGreen,backgroundcolor=OliveGreen!25,bordercolor=OliveGreen,#1]{#2}}
\newcommandx{\improvement}[2][1=]{\todo[linecolor=Plum,backgroundcolor=Plum!25,bordercolor=Plum,#1]{#2}}
\newcommandx{\thiswillnotshow}[2][1=]{\todo[disable,#1]{#2}}

\newcommand{\bs}{BosphorusSign\xspace}
\newcommand{\bsk}{BosphorusSign\textbf{22k}\xspace}

\title{\bf \bsk Sign Language Recognition Dataset}

\usepackage{authblk}

\name{Oğulcan Özdemir\textsuperscript{a}, Ahmet Alp Kındıroğlu\textsuperscript{a}, Necati Cihan Camgöz\textsuperscript{b}, Lale Akarun\textsuperscript{a}}

\address{\textsuperscript{a}Boğaziçi University, Computer Engineering Department, Istanbul, Turkey \\
         \textsuperscript{b}CVSSP, University of Surrey, Guildford, United Kingdom \\
         \{ogulcan.ozdemir, alp.kindiroglu, akarun\}@boun.edu.tr, n.camgoz@surrey.ac.uk\\}

\abstract{
Sign Language Recognition is a challenging research domain. It has recently seen several advancements with the increased availability of data. In this paper, we introduce the \bsk, a publicly available large scale sign language dataset aimed at computer vision, video recognition and deep learning research communities. The primary objective of this dataset is to serve as a new benchmark in Turkish Sign Language Recognition for its vast lexicon, the high number of repetitions by native signers, high recording quality, and the unique syntactic properties of the signs it encompasses. We also provide state-of-the-art human pose estimates to encourage other tasks such as Sign Language Production. We survey other publicly available datasets and expand on how \bsk can contribute to future research that is being made possible through the widespread availability of similar Sign Language resources. We have conducted extensive experiments and present baseline results to underpin future research on our dataset.
\\ \newline  
}

\begin{document}
\input{Misc/acronyms}

\maketitleabstract

\section{Introduction}
As native languages of the Deaf, Sign Languages (SL) are visio-temporal constructs which convey meaning through hand gestures, upper body motion, facial expressions and mouthings. \ac{aslr} is a challenging task and an active research field with the aim of reducing the dependency of sign language interpreters in the daily lives of the Deaf.

Among the many similar problems attempted by deep learning researchers, sign language recognition bears a resemblance to video-based action recognition because of its shared medium of information \cite{varol2017long}, and to speech recognition and machine translation problems \cite{bahar2019comparative,bahdanau2017learning}, due to its linguistic nature. However, there are certain aspects of \ac{aslr} that makes the task more challenging, one of which is the asynchronous multi-articulatory nature of the sign \cite{sutton1999linguistics}. Furthermore, the lack of large databases aimed at computer vision communities and the difficulty of annotating them has been an inhibiting factor in \ac{aslr} research \cite{hanke2010dgs,schembri2013building}.

In this paper, we present \bsk, an isolated SL dataset, for benchmarking repeatable deep learning experiments on SLR. The dataset was derived from \bs \cite{camgoz2016bosphorussign}, which has high-quality recordings collected from Deaf users of Turkish Sign Language (TID). The \bs Dataset was categorized linguistically, where sign glosses with the same meaning but a different set of morphemes were considered belonging to the same class. Although this annotation scheme was aimed to be useful in a Q\&A based interaction system, i.e., banking or hospital desk applications \cite{suzgun2015hospisign}, it is not well-suited for sign language recognition and production systems, where distinguishing between instances of similar sign classes with similar manual and non-manual features is essential. 

Although, \bs is publicly available, there are no benchmarks reported on this dataset. Furthermore, \bs does not have an evaluation protocol, making future research conducted using \bs dataset incomparable with one another.

Moreover, \bs dataset only provided skeleton information obtained from the Kinect V2 SDK. Although real-time depth-based skeleton estimation was state-of-the-art at the time of \bs's creation, it is jittery and lacks the crucial hand pose information, making the skeleton information provided in the \bs dataset inadequate for training human pose based sign recognition, translation and production models \cite{stoll2018sign,stoll2020text2sign}. 

To address these issues, in this paper, we have enhanced and refined the \bs dataset, to help future research in the fields of sign language recognition and production. The contributions of this work are listed as;
\begin{itemize}
    \vspace{-.05in}
    \item We have visually reviewed and cleaned up the dataset and removed all erroneous sign performances.
    \vspace{-.05in}
    \item We have revisited the labeling scheme and converted the linguistic categorization into a form more suitable for recognition and production research where each class shares the same manual features.
    \vspace{-.05in}
    \item We provide OpenPose \cite{cao2018openpose} body and finger coordinates in addition to Kinect V2 skeleton information.
    \vspace{-.05in}
    \item We have proposed an evaluation protocol and reported two benchmark results using 3D ResNets \cite{tran2018cvpr} and IDT \cite{wang2013action} to underpin future research on this dataset.
\end{itemize}

The rest of this paper is organized as follows: In Section~\ref{sec:related}, we give an overview of the SLR literature and other publicly available \ac{slr} datasets. In Section~\ref{sec:BOSPHORUSSIGN}, we introduce the new \bsk dataset. We then describe the baseline methods and share our experimental results in Section~\ref{sec:experiments}. Finally, we conclude this paper in Sections~\ref{sec:discuss} and \ref{sec:conclusion} by analyzing our baseline results, discussion and future work.

\section{Related Work}
\label{sec:related}

\begin{table*}[!ht]
\vspace{-0.08in}
\centering
\caption{Publicly available Isolated Sign Language Recognition datasets}
\begin{adjustbox}{width=1\linewidth}
\begin{tabular}{l|l|r|r|l|r|l|l}
\hline
Dataset              & Sign Language & \#Signers & Lexicon & Repetitions & \#Clips  & All Native Signers & Data Source \\ \hline
ASLLVD \cite{neidle2012asllvd} & American & 6 & 2,742 & arbitrary & 9,794 & Yes & RGB \\
Devisign \cite{chai2014devisign} & Chinese & 8       & 2,000    & 1-2         & 24,000  & No                 & Kinect v1      \\ 
BosphorusSign \cite{camgoz2016bosphorussign} & Turkish     & 6       & 855     & 4+          & 22,670  & Yes                & Kinect v2      \\ 
CSL \cite{zhang2016chinese} & Chinese & 50      & 500     & 5           & 125,000 & No                 & Kinect v2      \\ 
MS-ASL \cite{koller2019}    & American & 222     & 1,000    &  arbitrary           & 25,513  & Yes                & RGB            \\ 
\textbf{BosphorusSign22k} & \textbf{Turkish} & \textbf{6} & \textbf{744} & \textbf{4+} & \textbf{22,542} & \textbf{Yes} & \textbf{Kinect v2} \\
\hline
\end{tabular}
\end{adjustbox}
\label{tab:datasets}
\end{table*}

Since the work of \newcite{starner1998real}, there have been numerous studies on the isolated \ac{slr} task. More recently, utilization of state-of-the-art deep learning models \cite{koller2019weakly,zhang2016chinese} have resulted in better representation learning that is capable of achieving high accuracies over hundreds of unique sign glosses. Because of the spatio-temporal characteristics of the \ac{aslr} problem, popular methods from  video (human action \& activity recognition) \cite{wang2013action,carreira2017inflated} and speech recognition \cite{graves2005framewise} fields have been widely applied with success to the \ac{slr} problem. Since the focus of this paper is on computer vision based \ac{aslr} methods, the main variations among solutions proposed to this solution lie in their methods of data representation and temporal sign modeling.

Of the present methods in the literature, a large majority use sequences of RGB video frames and/or depth information \cite{cui2017recurrent,wang2016grassman,wang2018grassmann}. Some methods extract additional features from these input sources such as optical flow  and coordinates calculated by human pose capture methods such as Kinect \cite{shotton2012efficient} and OpenPose \cite{cao2018openpose}. A large number of methods extract state of the art features such as 2D \cite{koller2016deephand} and 3D \ac{cnn} outputs \cite{koller2019,huang2015sign,camgoz2016using}, \ac{idt} features \cite{ozdemir2016isolated}, hand appearance and trajectory features \cite{ozdemir2018isolated,he2016chinese,metaxas2018linguistically}. The use of spatial attention as in \newcite{yuan2019global} to focus learning on the signing space and temporal attention as in \newcite{camgoz2017subunets} and \newcite{camgoz2018neural,camgoz2020sign} are also among current popular research directions.

In terms of temporal segmentation and modeling, isolated videos of sign glosses often consist of varying length and complexity,  requiring the use of temporal modeling. In  \newcite{aran2008vision} and \newcite{zhang2016chinese}, \acp{hmm} are used with hand shape features and trajectories, while \newcite{koller2016deepsign} uses \acp{hmm} to train hand shape classifiers from weakly labeled sign videos. In \newcite{liu2016sign},  \acp{lstm}  are used with gradient histograms while in \newcite{camgoz2017subunets} they are used with \ac{ctc} and neural network features to learn sign languages. Based on the works of \newcite{koller2019} and \newcite{camgoz2016using} and the results of popular gesture recognition challenges such as Chalearn LAP \cite{wan2017results}, \ac{slr} studies with 3D \acp{cnn} currently tend to show higher performances in large datasets compared to other deep learning approaches utilizing \acp{lstm} and other popular methods. This generalization loses validity in the case of continuous \ac{slr} where the average clip length exceeds a few seconds.

In isolated Sign Language Recognition, the difficulties in obtaining high quality annotated videos has limited the amount of available public datasets. To the best of our knowledge, currently there exist four similar public available large scale isolated \ac{slr} datasets, which can be seen in Table~\ref{tab:datasets}, while Chinese Sign Language (CSL) recognition dataset \cite{zhang2016chinese} and MS-ASL \cite{koller2019} being the most recent ones. \bsk differentiates from these datasets as follows: Contrary to Chinese Sign Language (CSL) recognition dataset, which was recorded in front of a white background, \bsk dataset was captured using a Chroma Key background, which we believe will be beneficial for researchers who would like to utilize data augmentation techniques in their pipelines to improve their models generalization capabilities. We acknowledge the fact that MS-ASL is one of the new frontiers in sign language research as it initiated large scale ``in the wild''  isolated sign language recognition. However, the dataset is composed of publicly available YouTube videos. As Microsoft does not own the copyright of these videos, the availability of the data is not guaranteed. As of the submission of this paper, 290 videos are no longer publicly available which will potentially increase in the future, making comparison of future research against previously reported benchmarks harder. Additionally, there is the new SMILE DSGS corpus \cite{ebling2018smile}, which hasn't been fully publicly available and its evaluation protocol is yet to be defined.

\section{\bsk Dataset}
\label{sec:BOSPHORUSSIGN}
\begin{figure*}[t]
    \vspace{0.25in}   
    \begin{center}
       \includegraphics[width=0.90\linewidth]{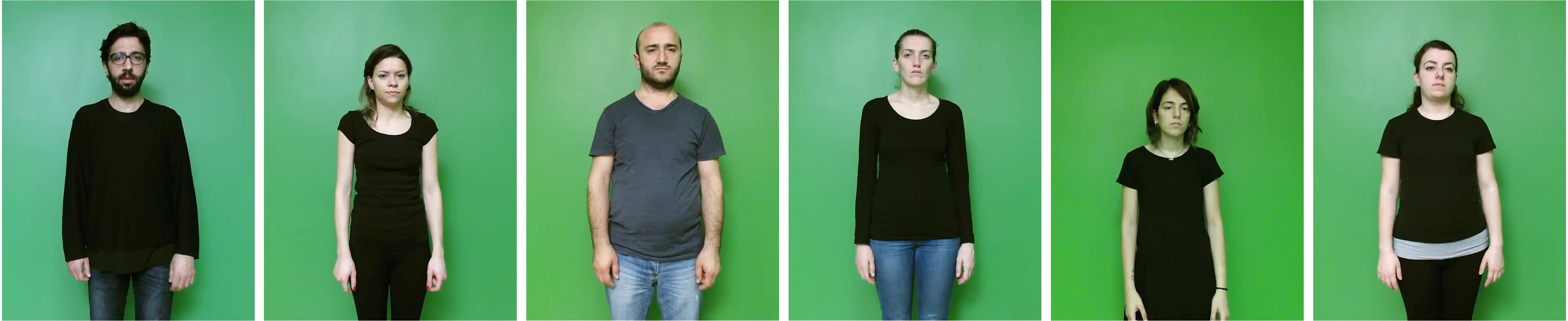}
    \end{center}
   \caption{Native signer participants of the \bsk dataset. \\(We propose using the left-most five signers as the training set and keep the remaining for evaluation.)}
   \label{fig:signers}
   \vspace{0.1in}
\end{figure*}

In this study, we present \bsk\footnote{\url{https://www.bosphorussign.com}}, a new benchmark dataset for vision-based user-independent isolated \ac{slr}. Our dataset is based on the \bs \cite{camgoz2016bosphorussign} corpus which was collected with the purpose of helping both linguistic and computer science communities. It contains isolated videos of Turkish Sign Language glosses from three different domains: Health, finance and commonly used everyday signs. Videos in this dataset were performed by six native signers, as shown in Figure \ref{fig:signers}, which makes this dataset valuable for user independent sign language studies.

All of the sign video recordings in the dataset were captured using Microsoft Kinect v2 \cite{zhang2012multimedia} with 1080p (1920x1080 pixels) video resolution at 30 frames per second. We believe having a higher resolution is essential for sign language recognition when interpreting the appearance information related to hand shape and movements. All of the videos share the same recording setup where signers stood in front of a Chroma-Key background which is 1.5 meter far away from the camera.

\begin{table}[!h]
  \centering
  \caption{Specifications of the \bsk dataset.}
  \begin{adjustbox}{width=1\linewidth}
    \begin{tabular}{l|l}
    \hline
        Property                                   & Description \\ \hline
        Number of sign classes                     & 744 \\ 
        Number of signers                          & 6   \\ 
        Number of videos                           & 22,542 \\ 
        Total Duration                             & $\sim$19 hours ($\sim$2M frames) \\
        \hline
        RGB Resolution                             &  1920 x 1080 pixels \\ 
        Depth Resolution                           &  512 x 424 pixels  \\ 
        Frame Rate                                 &  30 frames/second  \\
        \hline
        Body Pose Information (Kinect v2)          &  25 x 3D Keypoints \\ 
        Body Pose Information (OpenPose)           &  25 x 2D Keypoints \\
        Facial Landmarks (OpenPose)                &  70 x 2D Keypoints \\
        2 x Hand Pose Information (OpenPose)       &  21 x 2D Keypoints \\ 
        \hline
    \end{tabular}
    \end{adjustbox}
  \label{tab:specifications}
  \vspace{0.05in}
\end{table}

\begin{figure}[!b]
    \begin{center}
       \includegraphics[width=0.80\linewidth]{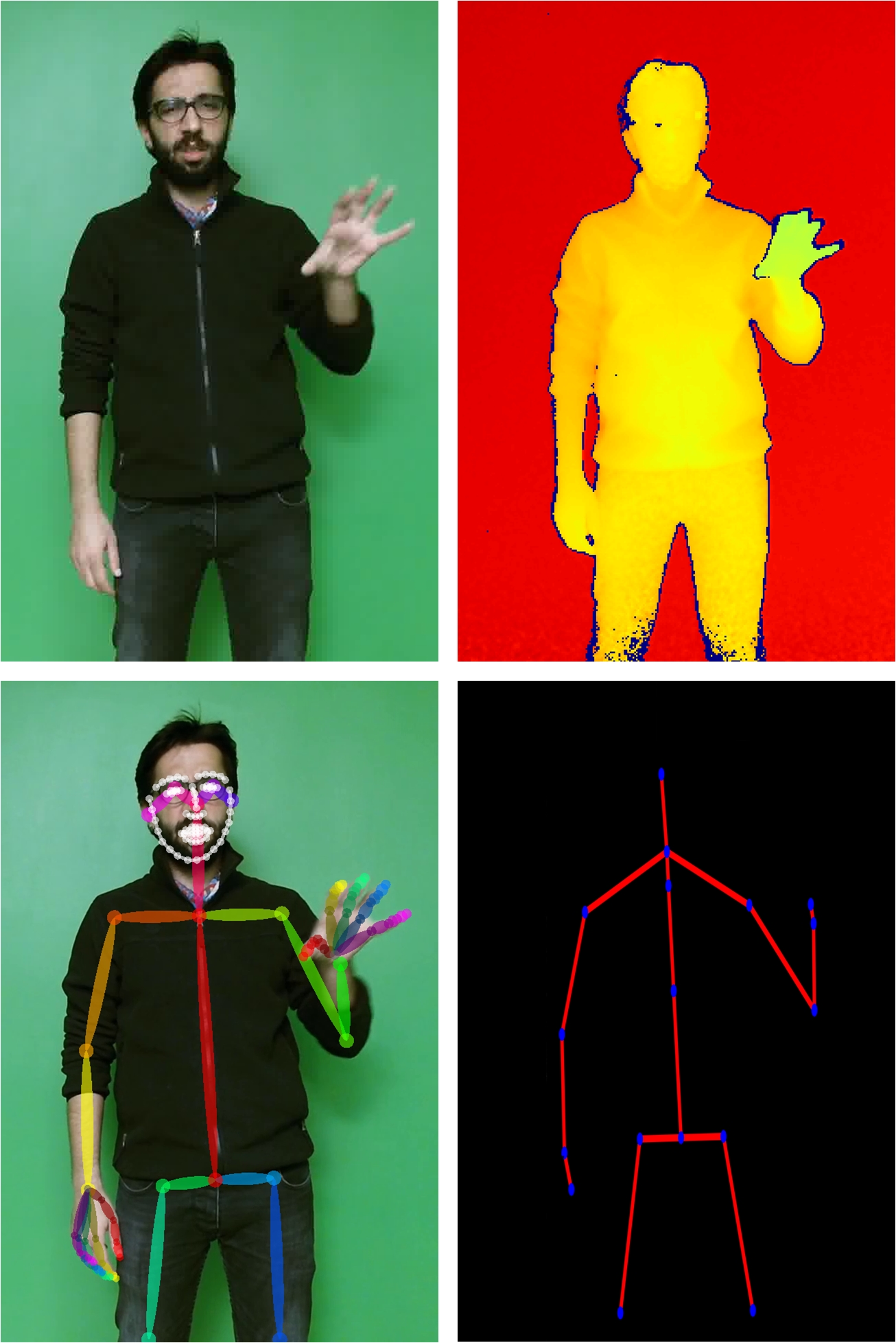}
    \end{center} 
   \caption{Modalities of \bsk. \\ Top: (left) RGB frame and (right) depth image \\ Bottom: (left) OpenPose and (right) Kinect v2 outputs.}
   \label{fig:sign-example}
\end{figure}

Specifications of the \bsk dataset can be seen in Table \ref{tab:specifications}. Since the dataset was collected using Microsoft Kinect v2, we provide RGB video, depth map and skeleton information of the signer for all sign videos in the dataset. Moreover, we also provide OpenPose \cite{cao2018openpose} joints, which include facial landmarks and hand joint positions in addition to body pose information. An example of provided modalities of the \bsk dataset can be seen in Figure \ref{fig:sign-example}.

\bsk has a vocabulary of 744 sign glosses; 428 in Health while having 163 in Finance domains as well as another 174 commonly used sign glosses. Properties of the proposed dataset and how it differentiates from the \bs corpus can be found in Table \ref{tab:comparison}.

\begin{table}[h]
\vspace{0.1in}
\captionsetup{justification=centering}
\caption{Properties of the publicly available subsets of the \bs corpus and the proposed \bsk datasets.}
\centering
\begin{adjustbox}{width=1\linewidth}
\begin{tabular}{l|c|c|c}
\hline
Dataset           & Lexicon & \# Clips & \# Repetitions \\ \hline
HospiSign \cite{camgoz2016sign}              & 33      & 1,257    & 6-8 \\
BosphorusSign \cite{camgoz2016bosphorussign} & 855     & --       & -- \\
\hspace{3mm}- Publicly Available             & 595     & 22,670   & 4+ \\
BosphorusSign22k                             & 744     & 22,542   & 4+ \\ 
\hspace{3mm}- General                        & 174     & 5,788    & 4+ \\
\hspace{3mm}- Finance                        & 163     & 4,998    & 4+ \\
\hspace{3mm}- Health                         & 428     & 11,756   & 4+ \\
\hline
\end{tabular}
\end{adjustbox}
\label{tab:comparison}
\vspace{0.1in}
\end{table}

In this work, we changed several aspects of the \bs dataset. First of all to set a baseline that would extend over the whole dataset, we merged all subsets and conducted our experiments accordingly. Further details of our experimental protocol can be found in the Section \ref{sec:protocol} Secondly, we manually inspected all the sign videos and eliminated erroneous recordings. Furthermore, we split signs that were semantically same but morphologically different. We also collapsed signs with similar manual features. The goal of this change was to benchmark the capabilities of the state-of-the-art models on learning meaningful representations for manual aspects of the sign glosses. However, we will be also releasing an uncollapsed version of the dataset. The changes on the \bs dataset are mostly focused on improving and cleaning the dataset and defining an evaluation protocol.  The dataset will be publicly available for research purposes upon submitting an EULA to the authors.

\section{Baseline Recognition Methods and Experiments}
\label{sec:experiments}
In this section, we provide training and test protocols of the \bsk dataset. We then describe baseline methods in detail and share our extensive experimental results.

\subsection{Experiment Protocol} 
\label{sec:protocol}
We defined our protocol by dividing the \bsk dataset into training and test sets in a signer independent manner where we use one signer for test and others for training. This yielded us a test set of 4,524 sign samples and a training set containing 18,018 samples. To set a baseline on the new dataset and the evaluation protocols, we perform isolated sign language recognition experiments and report classification accuracies on the test set with only one signer.

\subsection{Baseline Methods}
As for benchmark methods, there is no agreed-upon state-of-the-art approach in the isolated SLR literature, and researchers use different methods on different datasets \cite{camgoz2017subunets,koller2019,zhang2016chinese}. However, in the related field of action recognition, researches rely on benchmark datasets to compare their approaches against the state-of-the-art. Both 3D ResNets \cite{tran2018cvpr} and Improved Dense Trajectories - IDT \cite{wang2013action} are comparable state-of-the-art methods for action recognition, which have also yielded good performance on SLR \cite{ozdemir2016isolated}. Therefore, we have chosen 3D ResNets and IDT as our baseline approaches to cover both deep learning based representation learning techniques as well as hand crafted feature based methods.

\textbf{Improved Dense Trajectories - \ac{idt}:}
Although deep learning based models have become very popular recently, handcrafted approaches are still representative and competitive enough to be used in video recognition problems such as human action recognition and sign language recognition \cite{tran2018cvpr}. Inspired by this and to also give further insight to the reader instead of just reporting baselines using a deep learning based approach, we have used Improved Dense Trajectories (\ac{idt}) \cite{wang2013action} which is one of the most successful handcrafted methods with competitive performance for human action recognition and was used in sign language and gesture recognition recently \cite{ozdemir2016isolated,peng2014action}. \ac{idt} extracts local spatial features \ac{hog} \cite{dalal2005histograms}, and local temporal features \ac{hof} \cite{laptev2008learning} and \ac{mbh} \cite{dalal2006human} from the trajectories computed from dense optical flow field.

For recognizing sign language videos from the \bsk dataset, we have used a recognition pipeline similar to the one proposed by \newcite{wang2013action}. We first extracted trajectories from every sign video. After extracting trajectories, we randomly sampled trajectories from the training set, assuming these trajectories represent the overarching distribution. Then, \ac{pca} was applied to each component; namely \ac{hog}, \ac{hof} and \ac{mbh}. Using \ac{pca} outputs, we performed \ac{gmm} to cluster each component of the trajectories. Finally, \acp{fv} were computed from each component of trajectory descriptors from each sign video using the parameters of \ac{pca} and \ac{gmm}. Using these representations we trained a Linear \acp{svm} using different combinations of concatenated \acp{fv} components.

\textbf{3D Residual Networks with Mixed Convolutions: }
With the recent success of deep learning \cite{goodfellow2016deep} on tasks such as image and object recognition \cite{krizhevsky2012imagenet}, researchers have also started to build deep architectures for human action recognition where both spatial and temporal dimension are exploited \cite{simonyan2014two,tran2015learning,carreira2017inflated}. In this work, we have used a recently proposed video recognition method, which is based on 3D ResNet architecture with mixed 2D-3D convolutions, also called MC3 in \newcite{tran2018cvpr}. The model consist of two residual blocks with 3D convolutions, three residual blocks with 2D convolutions and a fully connected layer as its classification layer. This network was built based on the hypothesis that learning temporal dynamics is beneficial in early layers while the higher levels semantic knowledge can be learnt in late layers \cite{tran2018cvpr}. 

As our second baseline, we trained MC3 models and investigated the effects of fine-tuning different residual blocks of the network using the Kinetics-400 dataset \cite{carreira2017inflated}. The proposed training method for this model used randomized clips of frames as inputs, which is not suitable for our problem because randomized clips may include different non-recurring parts with the same isolated sign gloss. Therefore, at the training phase, we randomly sampled batches of uniformly sampled frames from sign videos to give our networks sufficient coverage over the frames.

\subsection{Implementation Details}
To evaluate baseline methods on the \bsk dataset, we have used the publicly available implementation in \newcite{wang2013action} for extracting \ac{idt} features and PyTorch \cite{paszke2017automatic} implementation of 3D ResNet model with mixed convolutions.

During training, we uniformly sampled 16 frames form each sign gloss with sizes of 112x112 pixels before feeding them to our networks. Sign clips are horizontally flipped a probability of 0.5 to be able to generalize over signers with different dominant hands. After preprocessing, we train our network using Adam optimizer \cite{diederik2014adam} with batch size of 32 on a NVIDIA Tesla V100 GPU. For testing, we performed the same preprocessing approach as in training except horizontal flipping of frame clips.

\subsection{Experimental Results} 
\label{sec:experiment}
We start our experiments by training several \ac{idt} based model with varying feature components, results of which can be seen in Table \ref{tab:idt}. Our results have shown that using motion features separately, \ac{hof} and \ac{mbh}, has yielded better results (83.29\% and 86.63\%) than using only appearance features, trajectory information (63.68\%) and \ac{hog} (76.59\%). Since the trajectory information only contains the position of the trajectory (mostly the position of hand regions in our case), it is expected that it cannot fully represent motion or appearance based features which have higher dimensionality and cannot encode more complex information about the sign or hand shape. 

\begin{table}[t]
  \centering
  \caption{Baseline \ac{idt} results on the \bsk dataset}
  \begin{adjustbox}{width=0.80\linewidth}
        \begin{tabular}{l|c}
        \hline
        Method            & Top-1 Acc (\%) \\ \hline
        TRAJ        & 63.68         \\
        \ac{hog}         & 76.59         \\
        \ac{hof}         & 83.29         \\
        \ac{mbh}         & 86.63         \\
        \ac{hog} + \ac{hof}     & 86.89         \\
        \ac{hog} + \ac{mbh}     & 86.98         \\
        \ac{hof} + \ac{mbh}     & 87.33         \\
        \ac{hog} + \ac{hof} + \ac{mbh} & \textbf{88.53}\\
        TRAJ + \ac{hog} + \ac{hof} + \ac{mbh}         &    87.86           \\ \hline
        \end{tabular}
    \end{adjustbox}
  \label{tab:idt}
\end{table}

Moreover, in the case of fusion of the trajectory components, our experiments have shown that using only \ac{hog}, \ac{hof} and \ac{mbh} features together improves our recognition accuracy (88.53\%), while adding the trajectory (TRAJ) information sightly decreases the performance of our system (87.86\%). Although the performance of our system is very close in the case of fusing multiple components, we can see that using \ac{hog} features with other features has improved our recognition performance in all cases. This supports the idea that appearance representation obtained using hand crafted features, such as \ac{hog}, is useful along with the temporal information when recognizing signs where the manual features of the sign are the main differentiating aspect between target classes.

As for our deep learning baseline, we performed several experiments on fine-tuning different residual blocks of the MC3 network \cite{tran2018cvpr}. In our first experiment, we first compared training our networks from scratch against using a pre-trained network (on Kinetics-400 dataset) and only training the final fully connected (fc) layer. 

\begin{table}[h]
  \centering
  \caption{Baseline 3D ResNet results on the \bsk dataset}
  \begin{adjustbox}{width=\linewidth}
        \begin{tabular}{l|c|c}
        \hline
        Method                      & Top-1 Acc (\%) & Top-5 Acc (\%) \\ \hline
        Training from scratch       & 57.76          & 84.22 \\
        Training only the last fc            & 55.03          & 81.98 \\ \hline
        Fine-tuning last 2 blocks    & 75.38          & 94.16 \\
        Fine-tuning last 3 blocks    & \textbf{78.85} & \textbf{94.76} \\
        Fine-tuning last 4 blocks    & 63.88          & 88.66 \\
        Fine-tuning all layers       & 71.02          & 92.51 \\ \hline
        \end{tabular}
    \end{adjustbox}
  \label{tab:resnet}
\end{table}

As it can be seen in Table \ref{tab:resnet}, training the network from scratch performed slightly better. We believe this is due to the fact that pre-trained network has never seen any sign samples, hence some of the essential spatio-temporal information that forms the sign is lost until it reaches the final fully connected layer. Using this insight, we decided to fine-tune other layers of the pre-trained network in additional to the last fully connected layer. We found that fine tuning the last 3 blocks to yield the best results for our task, an Top-1 accuracy of 78.85\% and and a Top-5 accuracy of 94.76\% on the test set.

\section{Analysis of Results and Discussion} \label{sec:discuss}

Although the 88.53\% Top-1 accuracy achieved by \ac{idt} is quite high for a 744 class problem, there is certainly still room for improvement. On the other hand, 3D ResNets, which are general-purpose video classification algorithms perform worse. We believe this is due to their inability to model longer-range temporal characteristics. Possible improvements include additional modalities and better temporal modeling.

\begin{figure}[h]
    \begin{center}
       \includegraphics[width=0.65\linewidth]{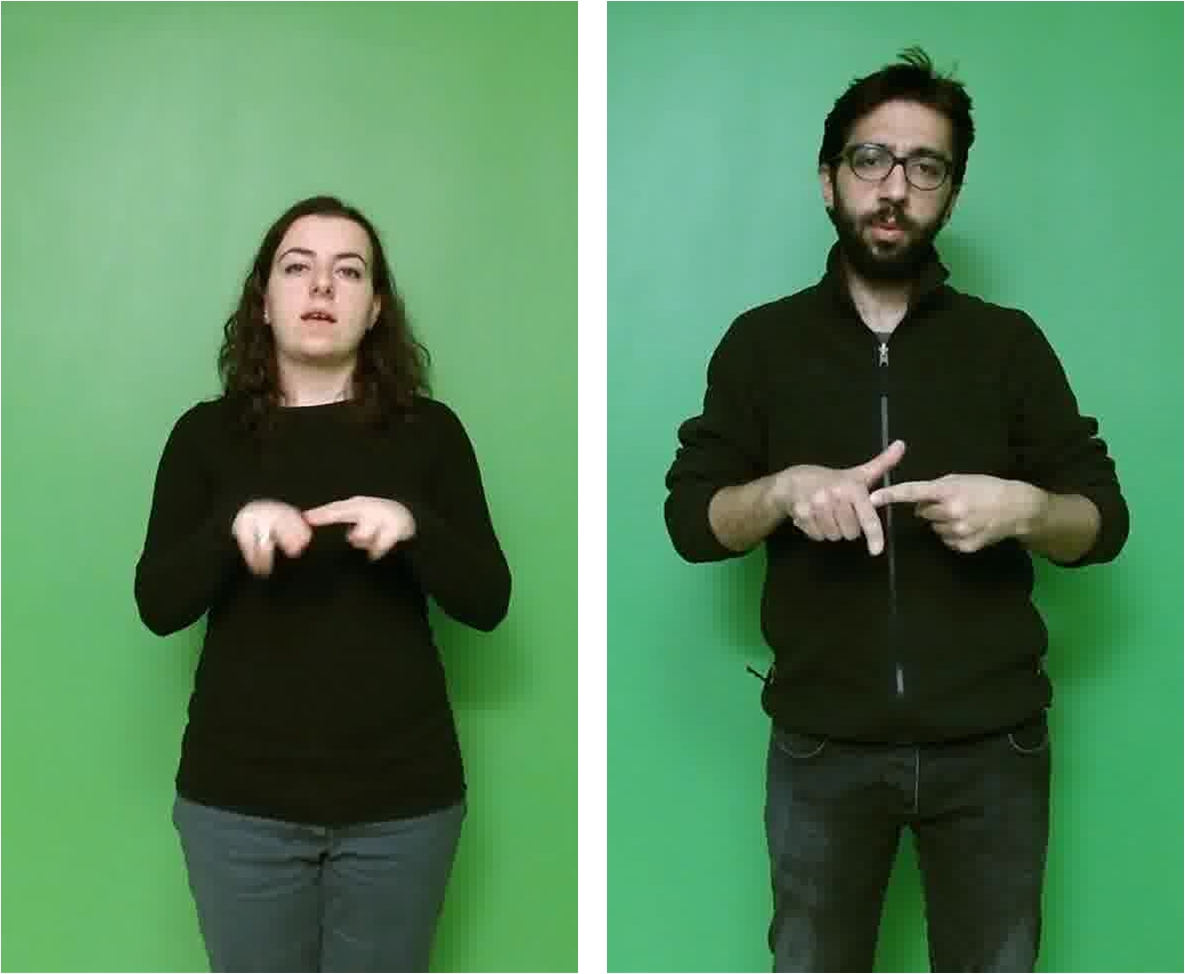}
    \end{center}
    \caption{Test sample of INSURANCE sign gloss (left) misclassified as FUND (right).}
    \label{fig:insurance-fund}
\end{figure}

We further investigated false classifications to gain further insight. For example, INSURANCE, INTERNET and COLD sign glosses are commonly confused with FUND, TEACHING and FACE respectively. Upon investigation we discovered that this is due to baseline methods' inability to model fine grained hand shapes. As seen in Figure \ref{fig:insurance-fund}, while our models were able to distinguish the sign by its motion, it fails to discriminate it using the similar handshapes. We believe this is due to the image resolution our networks are trained for in the case of MC3 model and the representation limitations of the \ac{hog} features for our \ac{idt} baseline. One way to tackle this problem could be to utilize specialized networks, such as Deep Hand proposed by \newcite{koller2016deephand}, and use it as another modality in our recognition pipeline.

\begin{figure}[t]
    \begin{center}
       \includegraphics[width=0.65\linewidth]{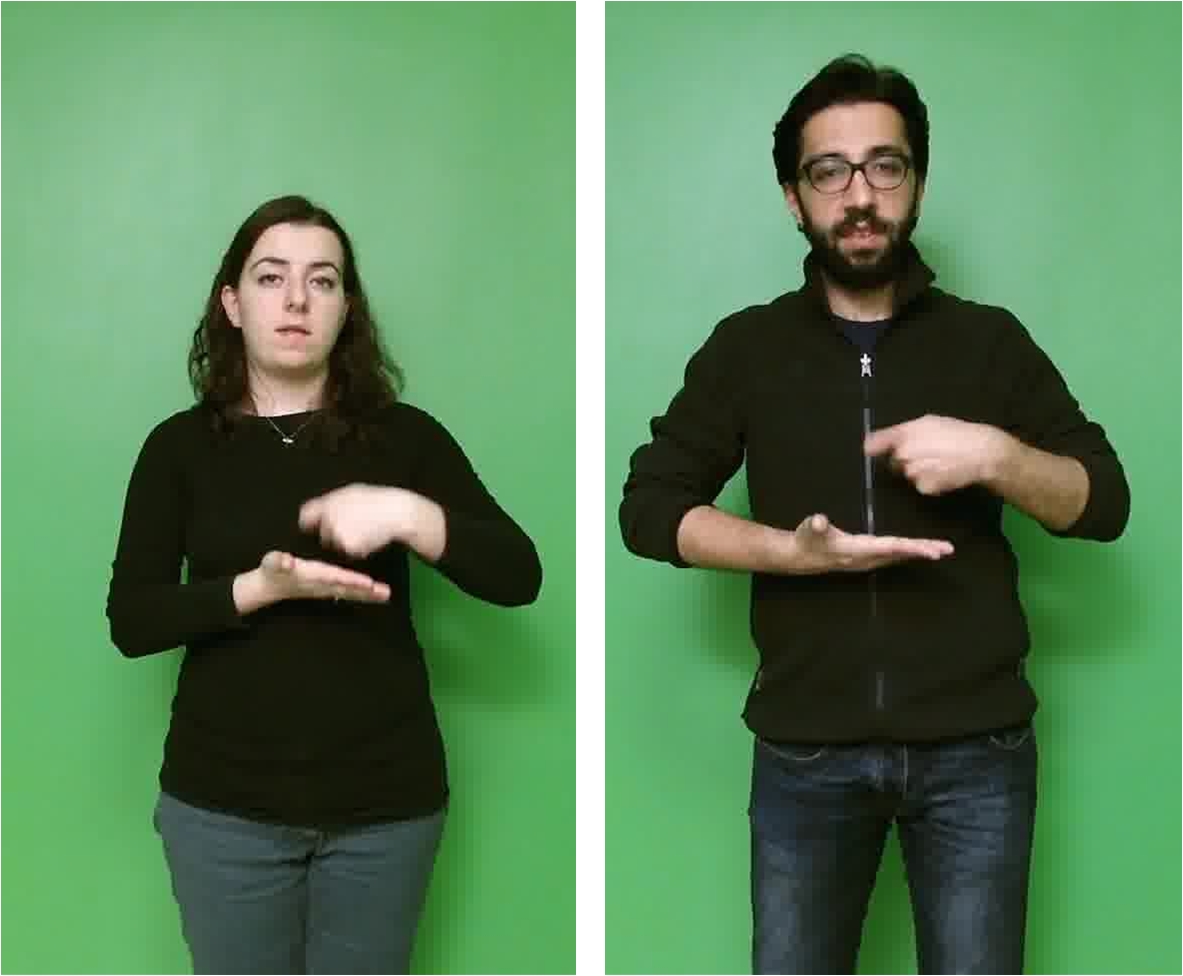}
    \end{center}
    \vspace{-0.1in}
    \caption{Test sample of PRICE sign (left) misclassified as SHOPPING (right)}
   \label{fig:price-shopping}
\end{figure}

In addition, our analysis have shown that PRICE sign gloss is confused with SHOPPING sign gloss (see Figure \ref{fig:price-shopping}) because the number of repetitions of the same motion sub-unit is different in both signs. Although both signs have the same hand shape and movements, signers performing the SHOPPING sign gloss repeat the same motion sub-unit more than PRICE.

Furthermore, when looking at the Top-5 recognition accuracy on experiments with 3D ResNets, we can see that most of the misclassified signs are successfully classified among the top-5. Thus, we believe that focusing on problems mentioned above will help us to improve recognition performance. Baseline results also show that \ac{idt}, as a handcrafted approach, is still performing well on \ac{slr} as it can comprehensively model appearance and motion information obtained from the signer in the frame compared to the 3D ResNet model trained without any specific guidance. Another factor contributing to this performance disparity is the input size which is 112x112 for MC3 networks and 640x360 for the \ac{idt}.  

\section{Conclusion} \label{sec:conclusion}
In this paper, we present \bsk, a new signer-independent \ac{slr} evaluation benchmark. The dataset contains over 22k samples of isolated videos, of 744 unique Turkish Sign glosses performed by six native signers. To underpin future research, we applied two successful video recognition methods from the literature, namely \ac{idt} and 3D ResNets (MC3). We share our quantitative results as well as qualitative samples, providing further insight to the reader. 

As shown by our experimental results, there is still room for improvement in signer-independent \ac{slr} for cases where the manual aspects of the sign subtlety differentiates from other classes. As future work we plan to exploit the capture setup of our dataset, namely its suitability for data augmentation, and extend our protocol to investigate environment independent \ac{slr}. \bsk also enables further research into using the depth information to explore multi-modal fusion approaches.

\section{Acknowledgements}
This work has been supported by the TUBITAK Project \#117E059 and TAM Project \#2007K120610 under the Turkish Ministry of Development.

\section{References}
\label{reference}

\bibliographystyle{lrec}
\bibliography{references,Bib/action,Bib/camgoz,Bib/ctc,Bib/deeplearning,Bib/generative,Bib/gesture,Bib/misc,Bib/nmt,Bib/pose,Bib/seq2seq,Bib/sign,Bib/speech}

\end{document}

%% file: Misc/acronyms.tex
\begin{acronym}[PHOENIX14\textbf{T} ]

\acrodefplural{slrt}[SLRTs]{Sign Language Recognition Transformers}
\acrodefplural{sltt}[SLTTs]{Sign Language Translation Transformers}

\acro{slrt}[SLRT]{Sign Language Recognition Transformer}
\acro{sltt}[SLTT]{Sign Language Translation Transformer}

\acrodefplural{rnn}[RNNs]{Recurrent Neural Networks}
\acrodefplural{cnn}[CNNs]{Convolutional Neural Networks}
\acrodefplural{hmm}[HMMs]{Hidden Markov Models}
\acrodefplural{gru}[GRUs]{Gated Recurrent Units}
\acrodefplural{crf}[CRFs]{Conditional Random Fields}
\acrodefplural{gan}[GANs]{Generative Adversarial Networks}
\acrodefplural{lcs}[LCSes]{Longest Common Subsequences}
\acrodefplural{gpu}[GPUs]{Graphic Processing Units}
\acrodefplural{fv}[FVs]{Fisher Vectors}
\acrodefplural{svm}[SVMs]{Support Vector Machines}
\acrodefplural{hmm}[HMMs]{Hidden Markov Models}
\acrodefplural{lstm}[LSTMs]{Long Short-Term Memory Networks}

\acro{aslr}[ASLR]{Automatic Sign Language Recognition}
\acro{bsl}[BSL]{British Sign Language}
\acro{bleu}[BLEU]{Bilingual Evaluation Understudy}
\acro{blstm}[BLSTM]{Bidirectional Long Short-Term Memory}
\acro{cnn}[CNN]{Convolutional Neural Network}
\acro{crf}[CRF]{Conditional Random Field}
\acro{cslr}[CSLR]{Continuous Sign Language Recognition}
\acro{ctc}[CTC]{Connectionist Temporal Classification}
\acro{dl}[DL]{Deep Learning}
\acro{dgs}[DGS]{German Sign Language - Deutsche Gebärdensprache}
\acro{dsgs}[DSGS]{Swiss German Sign Language - Deutschschweizer Gebärdensprache}
\acro{fc}[FC]{Fully Connected}
\acro{fv}[FV]{Fisher Vector}
\acro{gan}[GAN]{Generative Adversarial Network}
\acro{gpu}[GPU]{Graphics Processing Unit}
\acro{gru}[GRU]{Gated Recurrent Unit}
\acro{gmm}[GMM]{Gaussian Mixture Model}
\acro{hmm}[HMM]{Hidden Markov Model}
\acro{hog}[HOG]{Histograms of Oriented Gradients}
\acro{hof}[HOF]{Histograms of Optical Flow}
\acro{ip}[IP]{Inner Product}
\acro{isl}[ISL]{Irish Sign Language}
\acro{idt}[IDT]{Improved Dense Trajectory}
\acro{lcs}[LCS]{Longest Common Subsequence}
\acro{lstm}[LSTM]{Long Short-Term Memory}
\acro{mbh}[MBH]{Motion Boundary Histograms}
\acro{nmt}[NMT]{Neural Machine Translation}
\acro{ocr}[OCR]{Optical Character Recognition}
\acro{ph12}[PHOENIX12]{RWTH-PHOENIX-Weather-2012}
\acro{ph14}[PHOENIX14]{RWTH-PHOENIX-Weather-2014}
\acro{ph14t}[PHOENIX14\textbf{T}]{RWTH-PHOENIX-Weather-2014\textbf{T}}
\acro{pca}[PCA]{Principal Component Analysis}
\acro{relu}[RELU]{Rectified Linear Units}
\acro{rnn}[RNN]{Recurrent Neural Network}
\acro{rouge}[ROUGE]{Recall-Oriented Understudy for Gisting Evaluation}
\acro{sgd}[SGD]{Stochastic Gradient Descent}
\acro{sla}[SLA]{Sign Language Assessment}
\acro{slr}[SLR]{Sign Language Recognition}
\acro{slt}[SLT]{Sign Language Translation}
\acro{sift}[SIFT]{Scale Invariant Feature Transform}
\acro{surf}[SURF]{Speeded Up Robust Features}
\acro{svm}[SVM]{Support Vector Machine}
\acro{wer}[WER]{Word Error Rate}

\end{acronym}